\documentclass{article}

\PassOptionsToPackage{numbers}{natbib}
\usepackage[final,main]{Formatting_Instructions_For_NeurIPS_2026/neurips_2026}

\usepackage[utf8]{inputenc}
\usepackage[T1]{fontenc}
\usepackage{hyperref}
\usepackage{url}
\usepackage{booktabs}
\usepackage{amsfonts}
\usepackage{amsmath,amssymb,amsthm}
\usepackage{mathtools}
\usepackage{nicefrac}
\usepackage{microtype}
\usepackage{xcolor}
\usepackage{graphicx}
\usepackage{subcaption}
\usepackage{float}
\usepackage{multirow}
\usepackage{enumitem}
\usepackage{array}
\usepackage{makecell}
\usepackage{wrapfig}   
\usepackage{needspace}

\title{NEXUS: Neural Energy Fields for Physically Consistent Contact-Rich 3D Object Dynamics}

\newcommand{\corrauth}{\textsuperscript{\dag}}

\author{%
  Qizhen Ying\corrauth \\
  Department of Engineering Science \\
  University of Oxford \\
  \texttt{qizhen.ying@eng.ox.ac.uk}
  \And
  Guangming Wang \\
  Department of Engineering \\
  University of Cambridge \\
  \texttt{gw462@cam.ac.uk}
  \And
  Yangchen Pan \\
  Department of Engineering Science \\
  University of Oxford \\
  \texttt{yangchen.pan@eng.ox.ac.uk}
  \And
  Victor Adrian Prisacariu \\
  Department of Engineering Science \\
  University of Oxford \\
  \texttt{victor.prisacariu@eng.ox.ac.uk}
  \And
  Brian Sheil \\
  Department of Engineering \\
  University of Cambridge \\
  \texttt{bbs24@cam.ac.uk}
  \And
  Yixiong Jing\corrauth \\
  Department of Engineering \\
  University of Cambridge \\
  \texttt{yj401@cam.ac.uk}
}

\newcommand{\method}{NEXUS}
\newcommand{\R}{\mathbb{R}}

\newcommand{\norm}[1]{\left\lVert #1 \right\rVert}

\newcommand{\vF}{\mathbf{F}}
\newcommand{\vx}{\mathbf{x}}
\newcommand{\vv}{\mathbf{v}}
\newcommand{\vf}{\mathbf{f}}
\newcommand{\mG}{\mathcal{G}}
\newcommand{\mE}{\mathcal{E}}
\newcommand{\mO}{\mathcal{O}}
\newcommand{\vg}{\mathbf{g}}
\newcommand{\vn}{\mathbf{n}}
\newcommand{\vz}{\mathbf{z}}
\newcommand{\vh}{\mathbf{h}}
\newcommand{\moe}{\mathrm{o\text{-}e}}
\newcommand{\moo}{\mathrm{o\text{-}o}}
\newcommand{\ve}{\mathbf{e}}
\newcommand{\veta}{\boldsymbol{\eta}}

\begin{document}

\maketitle
\begingroup
\renewcommand{\thefootnote}{\dag}
\footnotetext{Corresponding authors: Yixiong Jing and Qizhen Ying.}
\endgroup

\begin{abstract}
Physics-grounded video generation requires controllable 3D object dynamics that remain physically consistent under contact, deformation, and external forcing. Existing trajectory-based methods provide physical control, but they often model isolated physical effects rather than deriving dynamics from a unified physical structure, making it difficult to compose conservative and non-conservative effects while preserving controllability in contact-rich 3D scenes.
We present \method, a neural energy-field framework for contact-rich 3D object dynamics. \method\ represents each object as a structural graph and constructs dynamic contact graphs for object--object and object--environment interactions. Inspired by Hamiltonian Neural Network (HNN) energy-based dynamics, \method\ formulates motion through scalar energy and dissipation terms rather than directly predicting states or accelerations. Conservative effects, such as gravity and elastic deformation, are composed as additive energy terms over the scene. To handle non-conservative effects beyond standard HNNs, \method\ learns dissipation functions for damping and impact-induced energy loss. Forces are derived by differentiating the energy and dissipation functions and rolled out with a multi-substep semi-implicit integrator.
\method\ provides a stable and controllable physics reasoning module for contact-rich 3D object dynamics, improving long-horizon accuracy over representative learned and physics-structured dynamics baselines across controlled rollouts with varying mechanical properties and physical-effect compositions. Our contributions are: (i) an HNN-inspired energy-dissipation dynamics formulation that unifies conservative and non-conservative scene effects; (ii) a graph-based 3D representation for contact-rich object dynamics, which improves long-horizon rollout accuracy over the baselines across different mechanical settings; and (iii) a trajectory-guided video generation study showing that physically consistent motion improves downstream physical plausibility while maintaining competitive visual quality.
Code is available at \url{https://github.com/StephenYing/NEXUS}.
\end{abstract}

\section{Introduction}

Recent diffusion and autoregressive models have greatly improved the visual quality of video generation~\citep{NEURIPS2020_4c5bcfec,2022arXiv221002747L,2024arXiv241105902X}, and large video/world models can synthesize compelling rollouts from language, images, or actions~\citep{openai2024sora,kanervisto2025world,2025arXiv250920328W}. However, visual realism does not guarantee physical realism. Generated objects can float, penetrate, rebound incorrectly, or ignore force and material controls, limiting use in downstream settings such as robotics where planning requires physically consistent rollouts~\citep{2026arXiv260103782H,2022arXiv220614176W}.

To improve physical fidelity, some methods learn motion priors from physically rich data~\citep{2025arXiv251115684M,2025arXiv251205564W,2026arXiv260103782H,2026arXiv260111087Z}. Another line of work obtains trajectories from simulators or learned dynamics modules, and uses them to guide rendering or video synthesis~\citep{11093799,2025arXiv251204221B,2025arXiv250921309Y,xie2024physgaussian,zhang2024physdreamer,jiang2025phystwin,2025arXiv250920358W,namekata2026go}. Such methods are promising because it separates physical motion generation from visual synthesis and provide explicit control over the generated video. However, existing methods often focus on isolated physical effects or simplified 2D trajectory prediction.

Existing 3D trajectory generators remain limited in representing complex 3D physical interactions. They are often designed for one object interacting with an environment, rather than scenes where multiple deformable objects collide, deform, and exchange momentum~\citep{2025arXiv250920358W}. They also tend to inject physical effects such as gravity or external control as separate conditioning signals~\citep{2025arXiv250920358W,le2025gravity}, instead of composing them within a shared physical formulation. Finally, direct state or acceleration prediction lacks an explicit conservation structure, so numerical and prediction errors can accumulate as energy drift over long rollouts. The central challenge is therefore not only to provide trajectory guidance, but to represent contact-rich 3D dynamics through a unified physical interface where conservative and non-conservative effects can be composed, differentiated, and rolled out stably.

HNNs provide a principled energy-based alternative by learning scalar functions whose gradients define dynamics~\citep{hamiltonianNN2019}. HNN improves long-horizon rollout by moving beyond unconstrained state prediction. However, standard HNNs mainly target conservative systems with fixed and low-dimensional states. Extensions introduce dissipation and external forcing~\citep{zhong2021extending,deng2025denoising}, but remain focused on constrained mechanical settings. Contact-rich 3D object dynamics require a scene-level formulation that jointly represents deformable geometry, changing contact topology, external control, and impact-induced energy loss.

\begin{figure}[t]
  \centering
  \includegraphics[width=\linewidth]{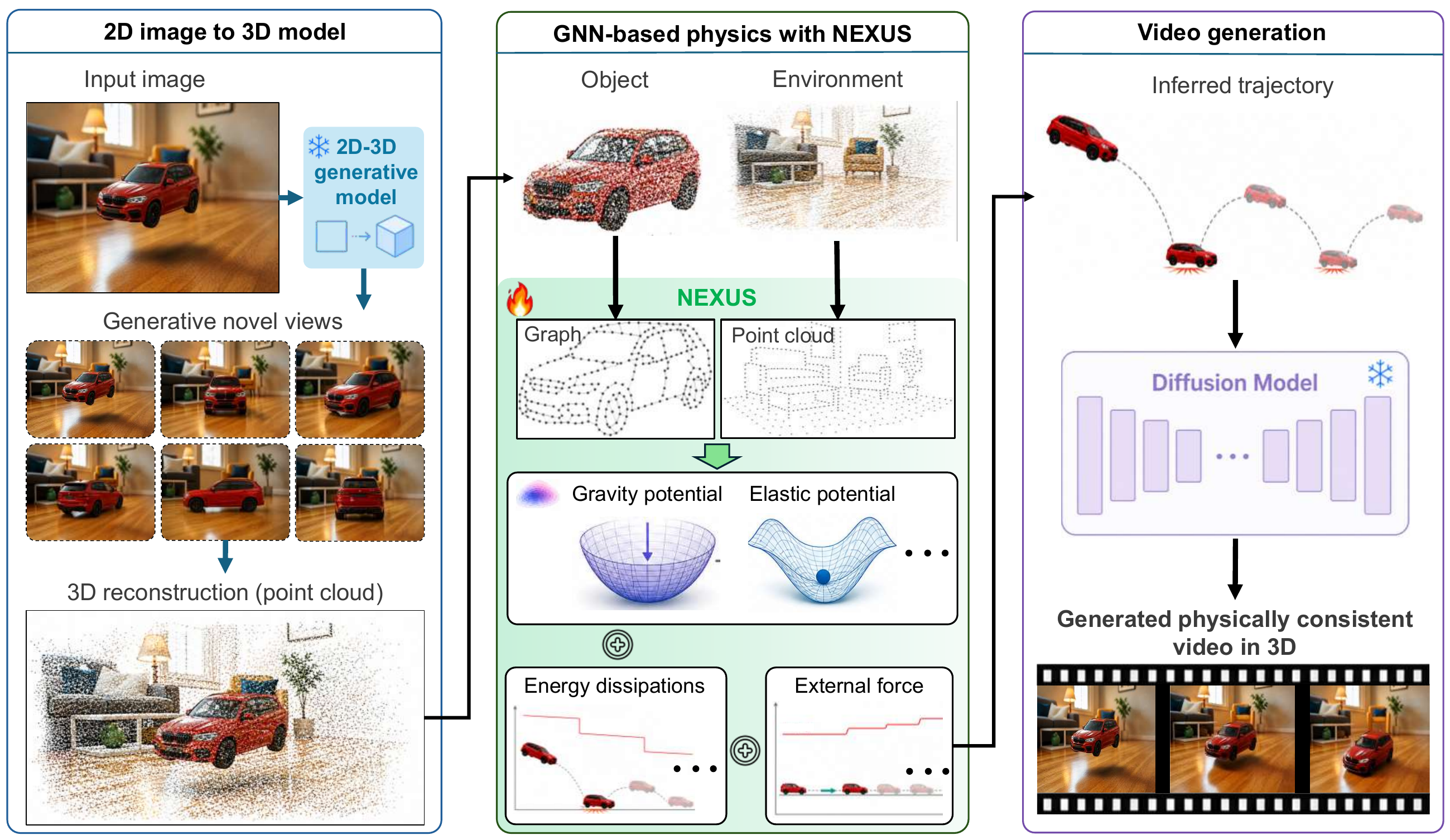}
  \caption{\textbf{\method\ overview.} A 2D-to-3D model reconstructs object and environment point clouds from an input image. \method{} predicts physically consistent trajectories through graph-based energy--dissipation reasoning. The inferred trajectory then guides a diffusion model for controllable physics-grounded video generation.}
  \label{fig:overview}
\end{figure}

To address these limitations, we propose NEXUS, a neural energy-field model for contact-rich 3D object dynamics (Figure~\ref{fig:overview}). NEXUS represents each object as a structural graph and builds dynamic object--environment and object--object contact graphs during rollout. It formulates conservative effects as additive scalar energy terms and models non-conservative impact loss with Rayleigh-style dissipation. Forces are obtained by differentiating the energy--dissipation functions and advanced with a multi-substep semi-implicit integrator. This turns contact-rich motion into energy exchange and dissipation under a shared scene-level dynamics interface, rather than independent object-wise trajectory prediction.


Our main contributions are following:
\begin{itemize}[leftmargin=2.5em,topsep=3pt,itemsep=2pt,parsep=0pt]
    \item \textbf{An energy--dissipation formulation for contact-rich 3D dynamics.}
    We derive forces from additive conservative energy terms and Rayleigh-style dissipation terms, giving a unified interface for structural elasticity, contact, external control, and impact-induced energy loss.
    \item \textbf{A graph-based point-cloud realization for multi-object interaction.}
    \method\ uses inferred structural graphs and dynamic object--environment/object--object contact graphs, with shared encoder and energy--dissipation heads across objects and contact types.
    \item \textbf{Controlled evidence for long-horizon physical consistency and video control.}
    \method\ improves long-horizon rollout accuracy across different mechanical settings and varied contact scene tests over the baselines, and video-generation studies validate the resulting physical control.
\end{itemize}

\section{Related Work}

\paragraph{Physics-grounded video generation with trajectory control.}
Trajectory-guided generation decouples motion generation from visual synthesis, enabling explicit control through learned or simulated motion~\citep{li2024generative,yang2025vlipp,wang2025wisa,zhang2025think,chefer2025videojam,pandey2025motion,feng2025physics,Yuan_2023_ICCV,2025arXiv250921309Y,lv2024gpt4motion,liu2024physgen,montanaro2024motioncraft,xie2025physanimator,2025arXiv251204221B,li2025wonderplay}. Recent 3D systems reconstruct or generate geometry and then simulate dynamics using MPM, spring--mass systems, or learned point-trajectory generators~\citep{li2023pacnerf,feng2024pie,xie2024physgaussian,zhong2024reconstruction,zhang2024dynamics,gao2025seeing,zhang2024physdreamer,jiang2025phystwin,2025arXiv250920358W}. These methods motivate explicit trajectory control, but the underlying dynamics are often simulator-dependent or learned as direct trajectories rather than as a compositional energy--dissipation model.

\paragraph{Neural physical dynamics.}
Learned simulators represent physical systems as objects, particles, meshes, or point clouds and predict state updates with neural networks. Interaction Networks~\citep{battaglia2016interaction}, graph simulators~\citep{li2018learning,sanchez2020learning,pfaff2020learning}, and recent geometric dynamics models~\citep{zhang2024dynamics,kim2024object} scale to high-dimensional states and complex geometry. However, most are trained as direct transition, acceleration, or displacement predictors. This flexibility comes without an explicit decomposition of conservative forces, contact, external work, and dissipation, making long-horizon stability and physical-effect composition difficult.

\paragraph{Energy-based Hamiltonian neural dynamics.} 
Hamiltonian and Lagrangian neural networks learn scalar mechanical functions and recover dynamics by differentiation, improving structure preservation in conservative systems~\citep{hamiltonianNN2019,cranmer2020lagrangian,zhong2019symplectic,finzi2020simplifying,deng2025denoising}. Dissipative and port-Hamiltonian variants introduce smooth energy loss~\citep{zhong2020dissipative,desai2021port,sosanya2022dissipative}, and contact-aware extensions adapt differentiable contact handling to low-dimensional multi-object systems~\citep{zhong2021extending}. NEXUS moves this energy-based view to contact-rich 3D point-cloud dynamics by combining structural graphs, dynamic contact graphs, additive potentials, and learned Rayleigh-style impact dissipation.

\section{Method}
\label{sec:method}

\subsection{Problem Setup}

\paragraph{Scene state and controls.}

\begin{figure}[t]
  \centering
  \includegraphics[width=\linewidth]{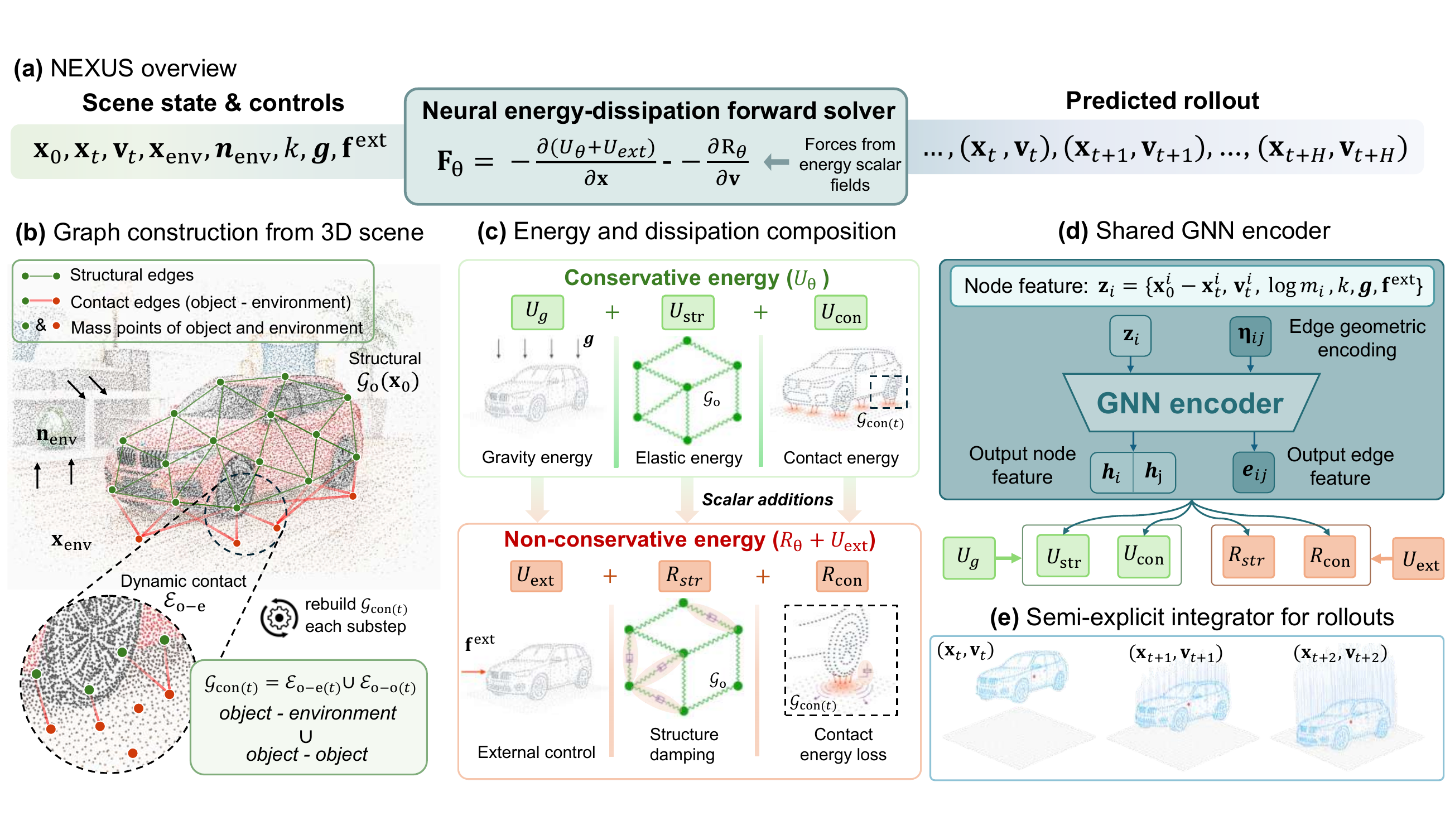}
  \caption{\textbf{\method\ architecture.}
  Overview of the \method{} architecture. (a) \method{} predicts rollouts from scene states and physical controls using an energy--dissipation forward solver. (b) Structural graphs represent object geometry, while dynamic contact graphs model object--environment and object--object interactions. (c) Conservative and non-conservative effects are represented as composable scalar terms. (d) A shared GNN encoder parameterizes the energy and dissipation modules. (e) Differentiated forces are advanced with a multi-substep semi-implicit integrator for rollout prediction.}
  \label{fig:architecture}
\end{figure}

We model a scene with $M$ dynamic objects.
Object $a$ contains $N_a$ particles, and all dynamic particles are concatenated into positions $\vx_t\in\R^{N\times 3}$ and velocities $\vv_t\in\R^{N\times 3}$, where $N=\sum_{a=1}^{M}N_a$ (as shown in Figure~\ref{fig:architecture}(a)).
The undeformed object geometry is $\vx_0$ at time $t=0$, with $\vx_0^i$ denoting the initial position of particle $i$.
The static environment is represented as a point cloud $\vx_{\mathrm{env}}\in\R^{N_e\times 3}$ with normals $\vn_{\mathrm{env}}\in\R^{N_e\times 3}$, where $N_e$ is the number of environment points.
Each dynamic particle has three properties, i.e., an object id, type attribute $\tau_i$, and mass $m_i$.
The physical inputs at time $t$ are stiffness $k$, gravity $\vg$, and masked external control/force $\vf^{\mathrm{ext}}_{i,t}$.
Given $(\vx_t,\vv_t,\vx_0,\vx_{\mathrm{env}},\vn_{\mathrm{env}},k,\vg,\vf^{\mathrm{ext}}_t)$ with $\vf^{\mathrm{ext}}_t\in\R^{N\times 3}$, the goal is to predict $\{(\vx_{t+h},\vv_{t+h})\}_{h=1}^{H}$ in closed loop for length $H$ as shown in Figure~\ref{fig:architecture}(a).

\paragraph{Energy/Rayleigh dynamics.}
\method{} represents objects and environment with structural and contact graphs, and evaluates conservative energy and Rayleigh-style dissipation terms whose derivatives define forces:
\begin{equation}
    \vF_\theta(\vx,\vv) = -\frac{\partial \left(U_\theta(\vx,\vv;k,\vg)+U_{\mathrm{ext}}(\vx;\vf^{\mathrm{ext}})\right)}{\partial \vx}
    - \frac{\partial R_\theta(\vx,\vv;k)}{\partial \vv}.
    \label{eq:force}
\end{equation}
The conservative energy terms $U_\theta$ compose gravity, structural elasticity, and contact potential in a single scalar interface.
The external-force input is represented by $U_{\mathrm{ext}}$ and combined with $R_\theta$ as the non-conservative part.
The dissipation terms $R_\theta$ follow a Rayleigh-style form and are differentiated with respect to velocity to produce non-conservative damping and impact energy loss.

\subsection{Graph Construction and Encoding}
\label{sec:graphs}

\paragraph{Shared encoder: node/edge inputs and outputs.}
We use a single shared graph encoder as shown in Figure~\ref{fig:architecture}(d).
For particle $i$, the node input is defined as:
\begin{equation}
    \vz_i=[\vx_i-\vx_0^i,\;\vv_i,\;\tau_i,\;\log m_i,\;\tilde{k},\;\vg,\;\vf^{\mathrm{ext}}_{i,t}],
    \label{eq:node-feature}
\end{equation}
where $\vz_i\in\R^{d_z}$ and $\tilde{k}$ is the normalized log-stiffness.
The encoder also receives local edge geometry $\veta_{ij}\in\R^{d_\eta}$ and produces shared node embeddings $\vh_i,\vh_j\in\R^{d_h}$ and edge embeddings $\ve_{ij}\in\R^{d_e}$.
For $\mE_0$, $\veta_{ij}$ includes relative displacement and rest length; for $\mE_{\moe}(t)$ and $\mE_{\moo}(t)$, it includes relative displacement, gap or distance, contact normal when available, relative normal velocity, and type indicators.
Additional GNN architecture choices are provided in Appendix~\ref{app:method-details}.

\paragraph{Structural graph.}
For each object, \method\ infers a time-invariant structural graph $\mG_0=(\mathcal{V}_{\mathrm{obj}},\mE_0,\ell^0)$ from $\vx_0$ (representing undeformed objects) as shown in Figure~\ref{fig:architecture}(b), where $\mathcal{V}_{\mathrm{obj}}$ and $\mE_0$ represent the nodes and edges of $\mG_0$. $\mE_0$ connects particles only within the same object and stores their rest length $\ell^0_{ij}=\norm{\vx_i^0-\vx_j^0}_2$.
$\mE_0$ encode internal elastic structure and $\mG_0$ remains unchanged at all rollout steps.
For multi-object scenes, \method{} uses one structural graph per object while sharing the same edge functions across objects.


\paragraph{Dynamic contact graph.}
At each time step $t$, \method\ defines a contact graph from $\vx_t$ (Figure~\ref{fig:architecture}(b)):
\begin{equation}
    \mG_{\mathrm{con}}(t)=(\mathcal{V}_{\mathrm{obj}}\cup\mathcal{V}_{\mathrm{env}},\mE_{\moe}(t)\cup\mE_{\moo}(t)).
\end{equation}
Object--environment edges $\mE_{\moe}(t)$ connect object particles to nearby environment points.
Object--object edges $\mE_{\moo}(t)$ are built for every unordered object pair ($a$ and $b$):
\begin{equation}
    \mE_{\moo}(t)=
    \bigcup_{a<b}
    \mathrm{kNN}(\vx_t^{(a)},\vx_t^{(b)})\cup
    \mathrm{kNN}(\vx_t^{(b)},\vx_t^{(a)}).
    \label{eq:oo-graph}
\end{equation}
The neighbor selection is conducted discretely at each step $t$. Thus, the selected topology is piecewise constant when deriving the force.
After the topology is selected, all contact features (e.g., gaps, distances, normals, and relative velocities between contact points) remain differentiable with respect to the continuous dynamic state.
The topology is rebuilt before each integration substep, allowing contact edges to appear, disappear, and move as the predicted geometry evolves.
Implementation settings for graph construction are listed in Appendix~\ref{app:method-details}.

\subsection{Energy Decomposition Intuition}
\label{sec:energy-dissipation}

Given the structural graph $\mE_0$ and contact graphs $\mE_{\moe}(t),\mE_{\moo}(t)$ defined above, \method{} evaluates scalar potential and Rayleigh-style dissipation terms on their edges.
Gradients are taken with respect to $\vx_t$ and $\vv_t$. $\vx_{\mathrm{env}}$, $\vn_{\mathrm{env}}$, $k$, $\vg$, and $\vf^{\mathrm{ext}}_{i,t}$ are treated as fixed inputs within each force evaluation.
This gives contact, deformation, external control, and dissipation a shared differentiable force interface.

\paragraph{Conservative energy terms.}
\method\ decomposes $U_\theta$ into different terms to account for varied physical effects, as shown in Figure~\ref{fig:architecture}(c):
\begin{equation}
    U_\theta(\vx,\vv;k,\vg)=
    U_{\mathrm{g}}(\vx;\vg)+U_{\mathrm{str},\theta}(\vx,\vv;k)+U_{\mathrm{con},\theta}(\vx,\vv;k),
    \label{eq:total-energy}
\end{equation}
where $U_{\mathrm{g}}$, $U_{\mathrm{str},\theta}$, and $U_{\mathrm{con},\theta}$ represent gravity, structural, and contact conservative terms (e.g., elastic spring energy), respectively. $U_{\mathrm{con},\theta}$ can be further decomposed into $U_{\moe,\theta}$ and $U_{\moo,\theta}$ for object--environment and object--object contact potentials.

The conservative gravity potential is
\begin{equation}
    U_{\mathrm{g}}(\vx;\vg)=
    -\sum_{i=1}^{N}\left(m_i \vg\right)^\top \vx_i,
    \label{eq:gravity-potential}
\end{equation}
so that $-\partial U_{\mathrm{g}}/\partial \vx_i=m_i \vg$.

For $(i,j)\in\mE_0$, the current length at $t$ is $r_{ij}=\norm{\vx_i-\vx_j}_2$, where $U_{\mathrm{str},\theta}$ is defined as a learned-coefficient spring energy:
\begin{equation}
    U_{\mathrm{str},\theta}=\sum_{(i,j)\in\mE_0}
    \frac{1}{2}k^{\mathrm{eff}}_{ij}(\ve_{ij},k)(r_{ij}-\ell^0_{ij})^2,
    \qquad k^{\mathrm{eff}}_{ij}>0.
    \label{eq:struct-energy}
\end{equation}
The coefficient $k^{\mathrm{eff}}_{ij}$ is predicted from the edge embedding $\ve_{ij}$ produced by the shared encoder and the physical stiffness input $k$, so it can use the current state $(\vx,\vv)$ when parameterizing structural elasticity.
For $(i,j) \in \{\mE_{\moe}(t) \cup \mE_{\moo}(t)\}$, NEXUS defines a penetration margin $\delta_{ij}$ from point-plane distance for object--environment contact and from radial point-point distance for object--object contact.
With $\phi(\delta)=\max(0,\delta)$, the contact potential is:
\begin{equation}
    U_{\mathrm{con},\theta}=\sum_{(i,j)\in\mE_{\moe}(t)\cup\mE_{\moo}(t)}
    w_{ij}\frac{1}{2}k^{\mathrm{con}}_{ij}(\ve_{ij},k)\phi(\delta_{ij})^2,
    \qquad k^{\mathrm{con}}_{ij}>0,
    \label{eq:contact-energy}
\end{equation}
where $w_{ij}$ is a fixed normalization weight for the contact edge, and  $k^{\mathrm{con}}_{ij}$ is a positive contact stiffness predicted from $\ve_{ij}$ and the physical stiffness input $k$, allowing contact response to depend on local geometry, contact type, and material stiffness.
The same definition of contact energy is used for $\{\mE_{\moe}(t) \cup \mE_{\moo}(t)\}$, with type-specific geometric features and weights $w_{ij}$.
For object--object contact, bidirectional $k$-NN edges are normalized so that the two directed queries represent one unordered pair interaction and do not double-count scalar contact contributions.

\paragraph{Non-conservative forcing and energy dissipation.}
The non-conservative external-force input is represented by
\begin{equation}
    U_{\mathrm{ext}}(\vx;\vf^{\mathrm{ext}})=
    -\sum_{i=1}^{N}\left(\vf^{\mathrm{ext}}_{i,t}\right)^\top \vx_i.
    \label{eq:external-work}
\end{equation}
Thus $-\partial U_{\mathrm{ext}}/\partial \vx_i=\vf^{\mathrm{ext}}_{i,t}$.

As shown in Figure~\ref{fig:architecture}(c), \method{} models dissipative non-conservative effects with two Rayleigh terms:
\begin{equation}
    R_\theta(\vx,\vv;k)=R_{\mathrm{str},\theta}(\vx,\vv;k)+R_{\mathrm{con},\theta}(\vx,\vv;k),
    \qquad R_{\mathrm{con},\theta}=R_{\moe,\theta}+R_{\moo,\theta}.
    \label{eq:dissipation}
\end{equation}
where $R_{\mathrm{str},\theta}$ represents damping along all edge pairs $(i,j)\in\mE_0$. $R_{\moe,\theta}$ and $R_{\moo,\theta}$ are contact dissipation terms on active object--environment and object--object contact edges, respectively.
Here $R_\theta$ is a Rayleigh dissipation function: a scalar function whose negative velocity gradient gives dissipative forces.
It is determined by the graph-conditioned edge embeddings together with geometry-dependent quantities, including structural edge directions, dynamic contact topology, penetration margins, contact normals, active-contact indicators, and the resulting damping coefficients.

Let $\hat d_{ij}(\vx)=(\vx_i-\vx_j)/(\norm{\vx_i-\vx_j}_2+\varepsilon)$ denote the current structural edge direction, and let
\begin{equation}
    \mathcal{A}_{\mathrm{con}}(\vx)
    =
    \{(i,j)\in \mE_{\moe}(t)\cup\mE_{\moo}(t):\phi(\delta_{ij}(\vx))>0\}
    \label{eq:active-contact-set}
\end{equation}
be the active contact set.
We then write the Rayleigh terms as
\begin{equation}
    \begin{aligned}
    R_{\mathrm{str},\theta}
    &=
    \sum_{(i,j)\in\mE_0}
    \frac{1}{2}c^{\mathrm{str}}_{ij}(\ve_{ij})
    \left((\vv_i-\vv_j)^\top\hat d_{ij}(\vx)\right)^2,\\
    R_{\mathrm{con},\theta}
    &=
    \sum_{(i,j)\in\mathcal{A}_{\mathrm{con}}(\vx)}
    \frac{w_{ij}}{2}c^{\mathrm{con}}_{ij}(\ve_{ij})
    \left[\mathrm{ReLU}(-\vv^n_{ij})\right]^2,
    \end{aligned}
    \qquad
    c^{\mathrm{str}}_{ij},c^{\mathrm{con}}_{ij}\ge 0.
    \label{eq:rayleigh-terms}
\end{equation}
where $c^{\mathrm{str}}_{ij}(\ve_{ij})$ and $c^{\mathrm{con}}_{ij}(\ve_{ij})$ are non-negative damping coefficients predicted from $\ve_{ij}$, and $\vv^n_{ij}=(\vv_i-\vv_j)^\top\vn_{ij}$ is the relative normal velocity, using zero environment velocity for object--environment contact.
The two squared velocity modes in Eq.~\eqref{eq:rayleigh-terms} correspond to structural stretch/compression along $\hat d_{ij}$ and approaching contact along $\vn_{ij}$.
Since damping forces use $-\partial R_\theta/\partial\vv$, these non-negative quadratic modes remove kinetic energy along those directions when the local coefficients are fixed.
The $\mathrm{ReLU}(-\vv^n_{ij})$ gate activates contact damping only for closing contacts, avoiding damping of tangential or separating motion.

\subsection{Integration and Closed-loop Training}
\label{sec:training}

Given $\vF_\theta$ from Eq.~\eqref{eq:force}, \method\ updates dynamic particles with $S$ semi-implicit substeps of size $\Delta t/S$, corresponding to the rollout module in Figure~\ref{fig:architecture}(e):
\begin{equation}
    \vv^{q+1}=\vv^q+\frac{\Delta t}{S}M^{-1}\vF_\theta(\vx^q,\vv^q),
    \qquad
    \vx^{q+1}=\vx^q+\frac{\Delta t}{S}\vv^{q+1},
    \label{eq:integrator}
\end{equation}
where the mass matrix is represented as $M=\mathrm{diag}(m_1,\ldots,m_N)\otimes I_3$.
Each numerical integration update in substep follows the same loop: rebuild $\mG_{\mathrm{con}}$ from the predicted state, encode nodes and edges, compute $U_\theta$ and $R_\theta$, differentiate them to obtain $\vF_\theta$, and integrate.
Thus contact topology is inferred from the model's own rollout during both training and evaluation.

The training objective is defined as a horizon-weighted rollout loss:
\begin{equation}
\mathcal{L}=
\sum_{h=1}^{H}\gamma^{h-1}
\left(
\lambda_x\ell_x^{(h)}+\lambda_v\ell_v^{(h)}
\right),
\label{eq:training-loss}
\end{equation}
where $\ell_x$ and $\ell_v$ are Huber losses on predicted positions and velocities.
We use a rollout curriculum from short to longer horizons; optimization details are provided in Appendix~\ref{app:baselines}.

\section{Experiments}
\label{sec:experiments}
We evaluate \method{} in three stages.
First, a controlled single-object benchmark compares long-horizon rollout accuracy under matched inputs, i.e., graphs, physical controls, and training budgets.
Second, to measure generalization to object--object contact, we evaluate \method{} on mixed-contact scenes with two, three, and five interacting objects.
Third, we use the predicted trajectories as physical guidance for contact-rich video generation.

\subsection{Experimental Setup}

\paragraph{Simulation benchmark.}

Our main quantitative benchmark contains a single dynamic object interacting with a static environment under $\vg$, masked $\vf_{ext}$, object--environment contact, and internal structural forces.
For each object and external-force schedule, we apply six different $k$ values (e.g., $k\in\{10,20,50,100,200,500\}$) to evaluate controlled material response.
We then evaluate mixed-contact generalization on two-, three-, and five-object scenes with both object--object and object--environment contact in Section~\ref{sec:multiobject-qual}.
Details regarding the dataset generation, particle counts, splits, and simulator settings are provided in Appendix~\ref{app:dataset}.

\paragraph{Metrics.}
The primary trajectory metric is rollout error at horizon $H$:
\begin{equation}
    \mathrm{RE@H}=\frac{1}{|\mO|}
    \sum_{i\in\mO}\norm{\hat{\vx}_{H,i}-\vx^{\mathrm{gt}}_{H,i}}_2 .
\end{equation}
For the single-object benchmark, we report $H\in\{1,6,12,24,48\}$ aggregated over all stiffness bins and evaluated episodes.
For multi-object evaluation, RE is endpoint error averaged over particles and episodes, not over time. Multi-object RE@48 measures 48 autoregressive updates against 49 ground-truth states, including the initial state.
We use long-horizon rollout error as the primary ranking metric, and report \emph{Strain} deformation error on the inferred structural graph, \emph{Max Pen.} environment penetration, and \emph{Speed} ratio $\norm{\hat{\vv}}/\norm{\vv^{\mathrm{gt}}}$ as complementary physical diagnostics.
For the speed ratio, values closer to $1$ are better.
All contact graphs and contact diagnostics during evaluation are computed from predicted states.

\paragraph{Baselines.}
We evaluate all baselines under the same experimental settings whenever the architectures can run within the available memory budget.
We first include three self-developed baselines: \emph{Explicit KNN Energy}, a physics-structured baseline with hand-specified structural and contact energies on the same inferred graph; \textit{Direct-State}, which predicts the next state directly; and \textit{Direct-Acceleration}, which predicts accelerations before integration.
To compare with representative neural simulators, we also include Interaction Network~\citep{battaglia2016interaction}, GNS~\citep{sanchez2020learning}, EGNN dynamics~\citep{satorras2021n}, and Point Transformer dynamics~\citep{zhao2021point}.
Details of the self-developed baselines are provided in Appendix~\ref{app:baselines}.
Single-object dynamics results are averaged over three random seeds, reported in Appendix Table~\ref{tab:seed-uncertainty}.

\subsection{Single-object Rollout Prediction}
\label{sec:single-object-results}

Table~\ref{tab:main-results} shows that \method\ achieves the lowest rollout error among all horizons (RE@1/6/12/24/48), while preserving low strain error and a near-unit speed ratio.
At RE@1, several direct-prediction and graph-based baselines are comparable, but their errors grow more quickly under closed-loop rollout.
Among the physics-structured baselines, Explicit KNN Energy benefits from an explicit contact prior but drifts at longer horizons because fixed-form energies cannot adapt to geometry-dependent contact and deformation.

\begin{table}[t]
\centering
\caption{Rollout prediction under matched inputs, controls, graphs, and training budgets. For the speed ratio, $1$ denotes perfect agreement, and values closer to $1$ are better.}
\label{tab:main-results}
\small
\setlength{\tabcolsep}{3.4pt}
\renewcommand{\arraystretch}{1.08}
\makebox[\textwidth][c]{\resizebox{1.04\textwidth}{!}{%
\begin{tabular}{@{}lcccccccc@{}}
\toprule
Model & RE@1 $\downarrow$ & RE@6 $\downarrow$ & RE@12 $\downarrow$ & RE@24 $\downarrow$ & RE@48 $\downarrow$ & Strain $\downarrow$ & Max Pen. $\downarrow$ & Speed $(\to 1)$ \\
\midrule
Explicit KNN Energy & 0.001 & 0.011 & 0.045 & 0.160 & 0.306 & 0.096 & 0.314 & 1.439 \\
Direct-State GNN & 0.003 & 0.018 & 0.040 & 0.135 & 0.231 & 0.157 & 0.032 & 0.662 \\
Direct-Acceleration GNN & 0.001 & 0.013 & 0.038 & 0.124 & 0.217 & 0.159 & \textbf{0.022} & 0.761 \\
\midrule
Interaction Network ~\citep{battaglia2016interaction} & 0.001 & 0.014 & 0.041 & 0.125 & 0.219 & 0.168 & 0.028 & 0.754 \\
Graph Network Simulator (GNS) ~\citep{sanchez2020learning} & 0.001 & 0.011 & 0.034 & 0.124 & 0.213 & 0.159 & 0.025 & 0.758 \\
EGNN Dynamics~\citep{satorras2021n} & 0.001 & 0.013 & 0.041 & 0.124 & 0.219 & 0.158 & 0.023 & 0.765 \\
Point Transformer Dynamics~\citep{zhao2021point} & 0.002 & 0.017 & 0.044 & 0.137 & 0.227 & 0.220 & 2.302 & 0.694 \\
\textbf{\method} & 0.001 & \textbf{0.005} & \textbf{0.022} & \textbf{0.087} & \textbf{0.137} & \textbf{0.066} & 0.145 & \textbf{1.020} \\
\bottomrule
\end{tabular}}}
\end{table}

\Needspace{0.2\textheight}
\subsection{Ablations and Controlled Behavior}
\label{sec:ablations-controlled}

\begin{wraptable}[14]{l}{0.50\textwidth}
\vspace{-8pt}
\centering
\caption{Ablations of \method{}.}
\label{tab:ablation-results}
\scriptsize
\setlength{\tabcolsep}{2.3pt}
\renewcommand{\arraystretch}{1.05}
\resizebox{\linewidth}{!}{%
\begin{tabular}{@{}lccccc@{}}
\toprule
Variant & RE@24 $\downarrow$ & RE@48 $\downarrow$ & Strain & \makecell{Max\\Pen.} $\downarrow$ & \makecell{Speed\\$(\to 1)$} \\
\midrule
Full \method{} & \textbf{0.087} & \textbf{0.137} & 0.066 & 0.145 & 1.020 \\
No contact & 1.683 & 7.349 & 0.059 & 7.868 & 5.743 \\
No dissipation & 0.095 & 0.154 & 0.084 & 0.130 & 1.137 \\
No structural energy & 0.235 & 0.353 & 1.151 & 0.077 & 1.100 \\
No stiffness $k$ & 0.190 & 0.177 & 0.091 & 0.087 & 1.055 \\
No extra force & 0.102 & 0.218 & 0.068 & 0.132 & 1.009 \\
Point-point contact & 0.864 & 3.666 & 0.077 & 7.511 & 3.112 \\
Local graph only & 0.114 & 0.173 & 0.069 & 0.154 & 1.040 \\
\bottomrule
\end{tabular}%
}
\vspace{-8pt}
\end{wraptable}

Table~\ref{tab:ablation-results} isolates the role of each component.
Removing contact, or replacing point-plane contact with point-point contact, causes catastrophic rollout failure.
Removing structural energy strongly worsens rollout and strain consistency.
Removing stiffness input hurts controlled generalization across $k$, and removing Rayleigh dissipation increases long-horizon error and yields a less realistic speed ratio.
The local-graph ablation further shows that non-local structure helps long-horizon deformation.
Together, these results support the full energy--dissipation decomposition rather than any single module alone.
Extra ablation regarding stiffness-conditioned behavior can be seen at Appendix~\ref{app:stiffness-results}.


\subsection{Multi-object Mixed-contact Generalization}
\label{sec:multiobject-qual}

\begin{table}[H]
\centering
\caption{Multi-object endpoint rollout error over 600 test episodes per object count; RE@48 uses 48 updates and lower is better. Strain@24 is time-averaged over steps 1--24. GNS and EGNN dynamics are omitted following out-of-memory failures on 24GB and 48GB GPUs in the original runs.}
\label{tab:multiobject-results}
\small
\setlength{\tabcolsep}{2.6pt}
\renewcommand{\arraystretch}{0.94}
\begin{tabular}{@{}llccccc@{}}
\toprule
Objects & Model & RE@6 $\downarrow$ & RE@12 $\downarrow$ & RE@24 $\downarrow$ & RE@48 $\downarrow$ & Strain@24 $\downarrow$ \\
\midrule
\multirow{6}{*}{2} & Explicit KNN Energy & 0.008 & 0.037 & 0.198 & 0.447 & 0.034 \\
 & Direct-State GNN & 0.012 & 0.037 & 0.236 & 0.429 & 0.077 \\
 & Direct-Acceleration GNN & 0.008 & 0.034 & 0.230 & 0.425 & 0.077 \\
 & Interaction Network~\citep{battaglia2016interaction} & 0.010 & 0.029 & 0.189 & 0.390 & 0.057 \\
 & Point Transformer Dynamics~\citep{zhao2021point} & 0.007 & 0.032 & 0.231 & 0.426 & 0.077 \\
 & \method{} & \textbf{0.005} & \textbf{0.022} & \textbf{0.175} & \textbf{0.378} & \textbf{0.030} \\
\midrule
\multirow{6}{*}{3} & Explicit KNN Energy & 0.009 & 0.054 & 0.214 & 0.470 & 0.036 \\
 & Direct-State GNN & 0.018 & 0.052 & 0.236 & 0.458 & 0.084 \\
 & Direct-Acceleration GNN & 0.012 & 0.047 & 0.234 & 0.450 & 0.080 \\
 & Interaction Network~\citep{battaglia2016interaction} & 0.008 & 0.038 & 0.203 & 0.430 & 0.062 \\
 & Point Transformer Dynamics~\citep{zhao2021point} & 0.011 & 0.047 & 0.234 & 0.449 & 0.080 \\
 & \method{} & \textbf{0.005} & \textbf{0.035} & \textbf{0.185} & \textbf{0.369} & \textbf{0.031} \\
\midrule
\multirow{6}{*}{5} & Explicit KNN Energy & 0.011 & 0.077 & 0.233 & 0.510 & 0.042 \\
 & Direct-State GNN & 0.013 & 0.080 & 0.256 & 0.493 & 0.093 \\
 & Direct-Acceleration GNN & 0.014 & 0.080 & 0.260 & 0.495 & 0.087 \\
 & Interaction Network~\citep{battaglia2016interaction} & 0.015 & 0.069 & 0.233 & 0.484 & 0.070 \\
 & Point Transformer Dynamics~\citep{zhao2021point} & 0.014 & 0.080 & 0.260 & 0.496 & 0.086 \\
 & \method{} & \textbf{0.007} & \textbf{0.059} & \textbf{0.194} & \textbf{0.406} & \textbf{0.035} \\
\bottomrule
\end{tabular}
\vspace{-6pt}
\end{table}

\begin{figure}[H]
  \centering
  \includegraphics[width=1.0\linewidth]{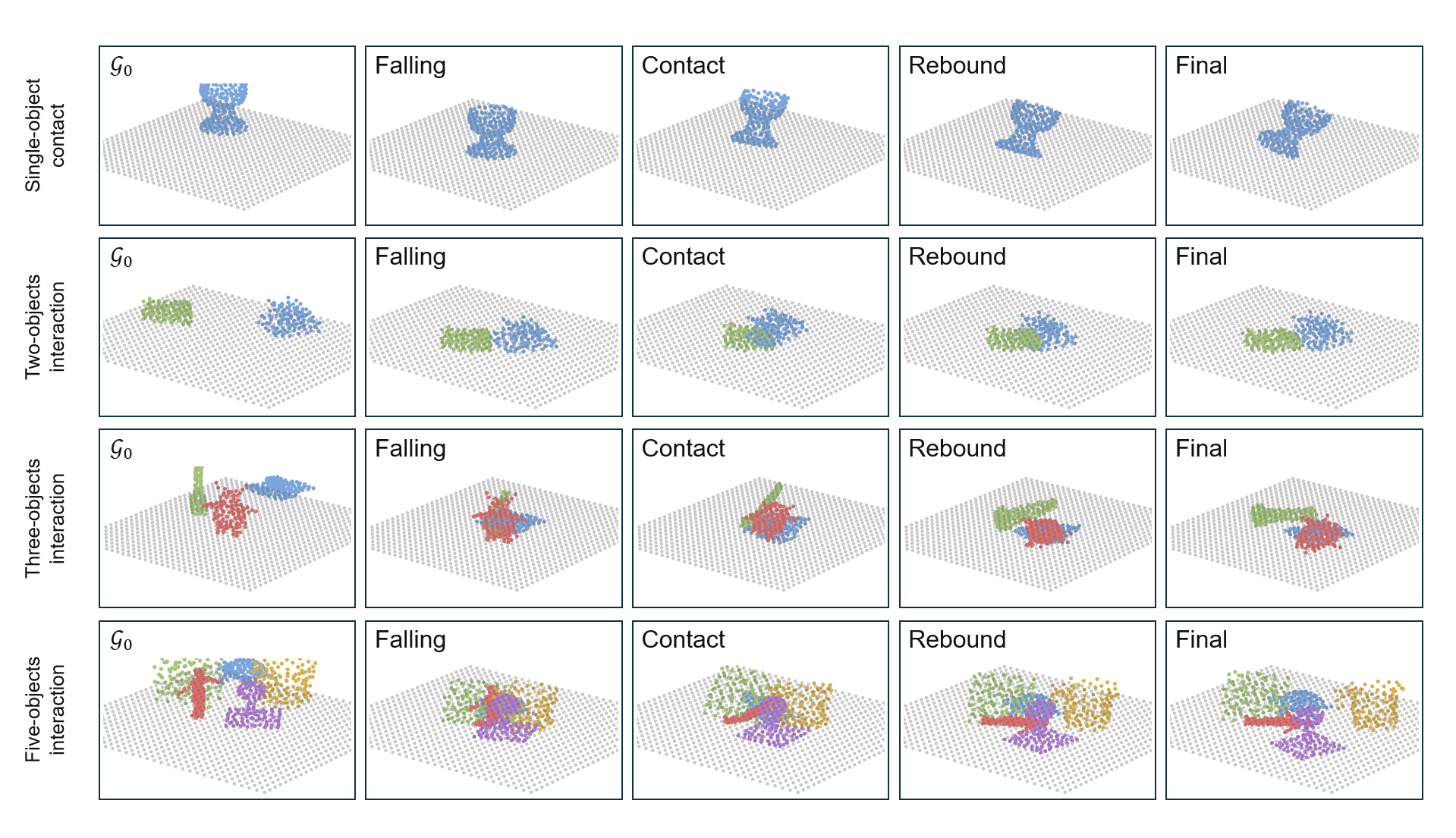}
    \caption{\textbf{Rollouts from single-object to mixed-contact scenes.}
        Representative top-view rollouts for one, two, three, and five dynamic objects over time in different stage. The increase of objects numbers brings more object--object interactions.}
  \label{fig:multiobject-qual}
\end{figure}

We further test the same formulation in mixed object--environment and object--object contact.
The compared methods share inferred per-object structural graphs and rebuild contact graphs from predicted states during closed-loop rollout.
Table~\ref{tab:multiobject-results} reports endpoint position errors for the six compared methods in two-, three-, and five-object scenes.
These experiments were run on a 24GB RTX 5090.
\method\ achieves the lowest endpoint error at every reported horizon and object count among the compared methods.
Figure~\ref{fig:multiobject-qual} visualizes representative top-view rollouts from the single-object reference case to two-, three-, and five-object mixed-contact scenes.

\subsection{Application: Video Generation}
\label{sec:video-generation}

Beyond trajectory prediction, we evaluate \method\ as a physics-grounded trajectory module for controllable video generation using the trajectory-conditioned video pipeline of PhysCtrl~\citep{2025arXiv250920358W}.
We keep the downstream image-to-video generator and dense 2D tracking-map interface fixed, and replace the PhysCtrl diffusion trajectory generator with \method.
Given an input image and physical controls, we reconstruct object/environment point clouds and use \method\ to predict 3D trajectories; these trajectories are projected to the image plane and used as dense tracking maps for the video generator.
Implementation details are provided in Appendix~\ref{app:video-details}.

\paragraph{Baselines.}
We compare with Wan2.2~\citep{wan2025}, CogVideoX-5B~\citep{yang2024cogvideox}, DragAnything~\citep{wu2024draganything}, ObjCtrl-2.5D~\citep{wang2024objctrl}, Veo3~\citep{googledeepmind2026veo}, and PhysCtrl~\citep{2025arXiv250920358W}.
Sparse-control methods receive a centroid or single-point trajectory according to their interface; \method\ uses dense projected multi-point trajectories.

\paragraph{Evaluation.}
We evaluate $6$ contact-rich test cases with GPT-based scoring and a blind human study with $10$ non-author raters.
Videos are scored on 5-point Semantic Adherence (SA), Physical Commonsense (PC), and Video Quality (VQ), using full videos as primary evidence with sampled frames as reference.
Prompts, evaluator instructions, and per-case settings are in Appendix~\ref{app:video-details}.

\begin{figure}[t]
  \centering
  \includegraphics[width=\linewidth]{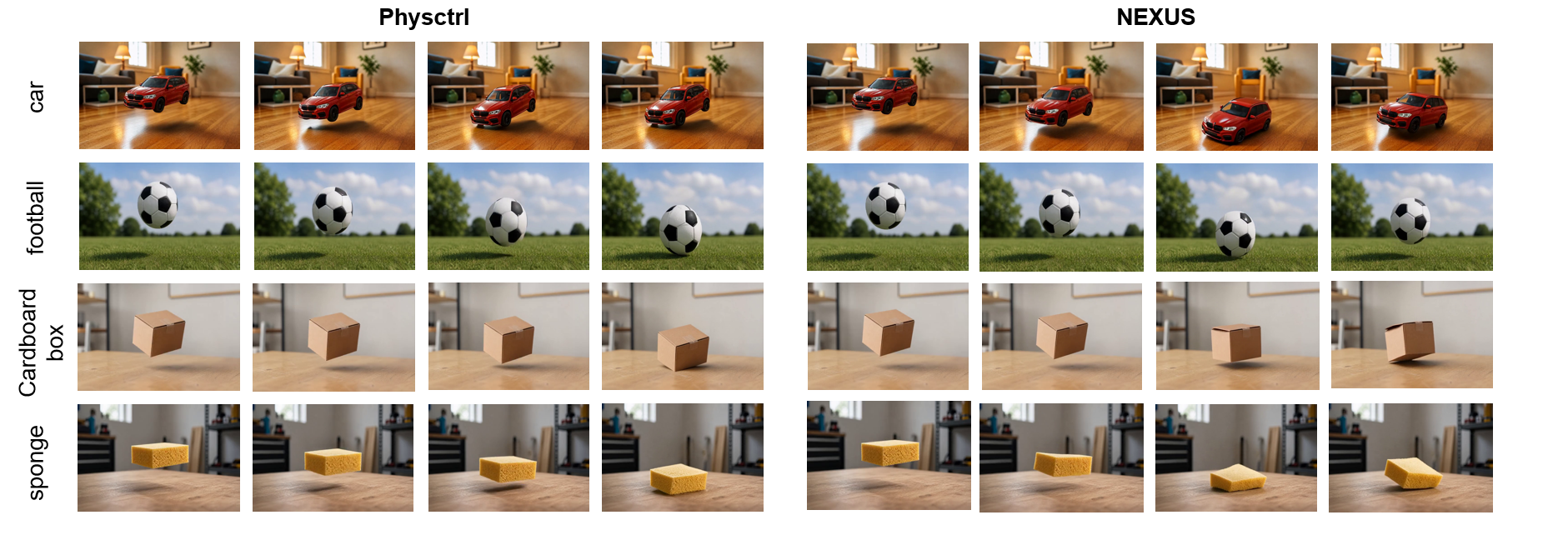}
  \caption{\textbf{Contact-rich video comparison with PhysCtrl.}
  \method\ trajectory guidance yields clearer collision, rebound, and settling under the same control intent.}
  \label{fig:video-qual}
\end{figure}

\paragraph{Results.}
Table~\ref{tab:video-eval} shows that \method\ achieves the best SA and PC while remaining competitive in VQ, and the blind human study follows the same trend and ranks \method\ highest on all three averages.
\Needspace{0.34\textheight}
\begin{wraptable}{l}{0.60\textwidth}
\vspace{-5pt}
\centering
\caption{Video generation evaluation.}
\label{tab:video-eval}
\scriptsize
\setlength{\tabcolsep}{2.4pt}
\renewcommand{\arraystretch}{1.05}
\resizebox{0.8\linewidth}{!}{%
\begin{tabular}{@{}lcccccc@{}}
\toprule
& \multicolumn{3}{c}{GPT} & \multicolumn{3}{c}{Human} \\
\cmidrule(lr){2-4}\cmidrule(lr){5-7}
Method 
& SA$\uparrow$ & PC$\uparrow$ & VQ$\uparrow$ 
& SA$\uparrow$ & PC$\uparrow$ & VQ$\uparrow$ \\
\midrule
Wan2.2 & 3.5 & 2.2 & 3.8 & 3.0 & 2.4 & 2.6 \\
CogVideoX-5B & 3.3 & 2.5 & 3.5 & 2.8 & 1.8 & 2.2 \\
DragAnything & 2.0 & 1.0 & 2.7 & 1.3 & 1.1 & 1.3 \\
ObjCtrl-2.5D & 2.3 & 1.5 & 3.3 & 1.8 & 1.5 & 1.3 \\
Veo3 & 3.3 & 2.3 & \textbf{4.3} & 2.6 & 2.0 & 2.3 \\
PhysCtrl & 3.0 & 2.7 & 4.0 & 2.6 & 1.8 & 2.0 \\
\method{} & \textbf{3.8} & \textbf{3.7} & 4.0 & \textbf{3.8} & \textbf{3.2} & \textbf{3.7} \\
\bottomrule
\end{tabular}%
}
\vspace{-5pt}
\end{wraptable}
Compared with text/image-only and sparse-control baselines, dense physics trajectories better preserve drop--contact--rebound--settle intent, reduce floating or identity-drift failures, and produce clearer post-impact damping.
Veo3 attains the strongest GPT-rated visual fidelity, but \method\ provides a stronger semantics--physics trade-off by substantially improving PC with high VQ. Figure~\ref{fig:video-qual} shows representative cases where \method-guided videos exhibit more explicit collision, rebound, and settling behavior than PhysCtrl~\cite{2025arXiv250920358W} under the same control settings.

\section{Discussion and Limitations}

\method\ demonstrates the value of an energy--dissipation structure for contact-rich point-cloud dynamics. The controlled single-object benchmark evaluates long-horizon accuracy, deformation, stiffness response, and contact behavior, while multi-object experiments demonstrate that the same architecture handles object--object and object--environment contact across different scene cardinalities. Larger and more diverse multi-object benchmarks are still needed. Separating trajectory prediction from visual synthesis also makes dynamics errors directly measurable and allows the physics module to be reused across video generators.


However the current setting uses simplified contact with limited frictional effects. Future work should examine richer contact metrics, explicit friction and rolling contact, larger multi-object benchmarks, higher-resolution particles, and broader video-generation studies.



\section{Conclusion}

We introduced \method, a neural energy-field framework for contact-rich 3D object dynamics. It combines energy and learned dissipation on structural and dynamic contact graphs, deriving forces by differentiation and advancing motion through numerical integration. Controlled benchmarks demonstrate improved long-horizon prediction over representative learned and physics-structured baselines. As a downstream application, \method\ trajectories guide video generation toward more physically plausible motion while maintaining competitive visual quality.





\begin{ack}
Yangchen Pan acknowledges the support from the Engineering and Physical Sciences Research Council (EPSRC) New Investigator Award under grant reference UKRI2775.

The authors acknowledge the use of resources provided by the Isambard-AI National AI Research Resource (AIRR)~\citep{mcintoshsmith2024isambardai}. Isambard-AI is operated by the University of Bristol and is funded by the UK Government's Department for Science, Innovation and Technology (DSIT) via UK Research and Innovation; and the Science and Technology Facilities Council [ST/AIRR/I-A-I/1023].
\end{ack}

{
\small
\bibliographystyle{unsrt}
\bibliography{references_v2}
}

\newpage
\appendix

\section{Additional Implementation Details}
\label{app:implementation}

\subsection{Dataset and Simulation Benchmark}
\label{app:dataset}

We construct a controlled point-cloud dynamics benchmark by running physics simulation on high-quality 3D objects curated from ObjaverseXL~\citep{deitke2023objaverse}.
Each episode contains object particles, an environment point cloud, per-frame object states, and control signals.
The state at time $t$ consists of particle positions $\vx_t\in\mathbb{R}^{N\times 3}$ and velocities $\vv_t\in\mathbb{R}^{N\times 3}$.
The single-object dataset contains $587$ unique objects and $35{,}172$ simulated trajectories.
Each single-object sample uses $256$ object particles and $1024$ environment/floor particles, resulting in $1280$ particles in total.
Each single-object trajectory has $48$ frames with a time step of approximately $0.0417$ seconds, corresponding to about $24$ FPS.
The evaluated stiffness values are $k\in\{10,20,50,100,200,500\}$.

\paragraph{Scene composition.}
The main benchmark contains object--environment contact with floor and environment geometry; multi-object contact is reported separately in Section~\ref{sec:multiobject-qual}.
For each episode, the object geometry is represented by a point cloud and the environment is represented by points with normals for contact evaluation.
The same observable inputs are provided to all dynamics models.

\paragraph{Controls.}
Each rollout is conditioned on gravity, stiffness, and optional masked external forces.
External forces are applied to selected object particles or regions according to the force mask.
For each object and force schedule, we vary stiffness to evaluate material-conditioned dynamics.

\paragraph{Splits and episode counts.}
For the single-object benchmark, we use a held-out object split based on \texttt{shape\_index}, with $90\%$ of objects used for training and $10\%$ used for validation/testing.
The training split contains $533$ objects and $31{,}932$ trajectories, while the validation/test split contains $54$ objects and $3{,}240$ trajectories.
Across the six stiffness values, this corresponds to $5{,}322$ training trajectories and $540$ validation/test trajectories per stiffness value.
Each full single-object trajectory contains $48$ frames.
During training, we use a rollout curriculum with rollout windows up to $24$ steps; evaluation reports horizons $H\in\{1,6,12,24,48\}$.

\paragraph{Multi-object evaluation data.}
For each object count, all methods are evaluated on 600 held-out episodes with 49 ground-truth states for 48 prediction updates. The trajectories are extended by one interval using the original simulator with 64 substeps, preserving the stored states and control schedules. The extension is validated by bitwise replay of the preceding interval.

\subsection{NEXUS Implementation Details}
\label{app:method-details}

\paragraph{Shared GNN encoder architecture.}
All learned scalar heads use the shared node/edge embeddings from a message-passing GNN encoder, with hidden width $128$, MLP depth $2$, and $4$ message-passing steps unless otherwise stated.
Structural/contact energy heads and Rayleigh heads are separate MLPs on top of the shared embeddings, while positivity constraints are enforced by softplus or exponential parameterizations as described below.

\paragraph{Structural graph preset.}
The default solid-graph preset builds object-internal local kNN edges with $k_{\mathrm{obj}}=18$, adds a minimum-spanning-tree connection to avoid disconnected components, and augments the graph with non-local geometric edges.
The non-local edges use $k_{\mathrm{long}}=8$ evenly selected neighbors from rank $12$ to $96$ in the initial point cloud.
Structural edges never connect particles from different dynamic objects.

\paragraph{Contact graph preset.}
Object--environment contact connects each object particle to $k_{\mathrm{con}}=8$ nearest environment points.
When environment normals are available, the signed point-plane gap for object particle $i$ and environment point $e$ is $g_{ie}=(\vx_i-\vx_e)^\top \vn_e$; otherwise we use point-point contact.
For object--object contact, each particle selects $k_{\moo}=4$ nearest particles from the other object in each pair, in both directions.
Object--object contact uses normal $\vn_{ij}=(\vx_i-\vx_j)/(\norm{\vx_i-\vx_j}_2+\varepsilon)$ and penetration margin $\delta_{ij}=2r_p\alpha_{\moo}-\norm{\vx_i-\vx_j}_2$, where $r_p$ is the particle contact radius and $\alpha_{\moo}=1$ by default.
We use mean-$k$ edge normalization; bidirectional object--object edges receive an additional factor of $0.5$ to avoid double counting in both contact energy and contact Rayleigh terms.

\paragraph{Positive coefficients.}
Learned stiffness and damping coefficients are constrained to be non-negative by softplus or exponential parameterizations.
For structural stiffness, we scale the positive coefficient by the user stiffness $k$ and a rest-length factor so that the same network can represent different material controls.
Coefficient scales are held fixed across ablations unless explicitly varied.

\subsection{Baseline Comparison Settings}
\label{app:baselines}

The comparison is designed to isolate the effect of the dynamics parameterization rather than differences in input information.
The comparison is designed to isolate the effect of the dynamics parameterization rather than differences in input information.

\paragraph{Matched inputs.}
All graph-based dynamics models receive object positions, velocities, stiffness control, gravity/external-force controls, and environment/contact information.
Contact graphs used at evaluation time are built from predicted states so that errors accumulated during rollout affect subsequent contact reasoning for all methods.

\paragraph{Rollout protocol.}
Models are evaluated autoregressively under matched rollout settings.
The trajectory metric is endpoint RE@H, reported at $H\in\{1,6,12,24,48\}$ for the single-object benchmark and $H\in\{6,12,24,48\}$ for multi-object scenes. Multi-object evaluation requires $49$ ground-truth states for $48$ updates, without horizon clipping.
All methods are evaluated across the same stiffness bins and test episodes.

\paragraph{Baseline families.}
We compare three groups of baselines.
First, we include \textbf{direct predictors}: \textit{Direct-State}, which predicts next-state transitions directly, and \textit{Direct-Acceleration}, which predicts accelerations and integrates forward; these test unconstrained transition and acceleration-regression parameterizations.
Second, we include \textbf{standard neural simulators}---Interaction Network~\citep{battaglia2016interaction}, Graph Network Simulator (GNS)~\citep{sanchez2020learning}, EGNN dynamics~\citep{satorras2021n}, and Point Transformer dynamics~\citep{zhao2021point}---to test whether generic message passing and geometric point-cloud inductive biases are sufficient without our energy--dissipation decomposition.
Third, we include a \textbf{physics-structured baseline}: \emph{Explicit KNN Energy}, which uses the same inferred structural/contact graphs but hand-specified structural/contact energy forms instead of geometry-conditioned neural scalar potentials.
This baseline isolates whether fixed-form physical approximations can account for long-horizon performance without fully learned scalar energy parameterization.

\paragraph{Training details.}
For the single-object benchmark, all learned dynamics models are trained with AdamW using a learning rate of $10^{-4}$ and weight decay of $10^{-6}$.
We use a ReduceLROnPlateau scheduler with decay factor $0.5$.
Models are trained for $20$ epochs, with $3072$ rollout windows sampled per training epoch and $256$ rollout windows sampled per validation epoch.
The batch size is one trajectory/window sample because graph sizes and rollout windows are handled with custom collation.
We use Huber loss with delta $1.0$, gradient clipping at $0.5$, and no state noise for either positions or velocities.
Training uses closed-loop rollout supervision with a curriculum: epochs $0$--$3$ use $6$ rollout steps, epochs $4$--$11$ use $12$ rollout steps, and epochs $12$--$20$ use $24$ rollout steps.
Single-object experiments were run on an NVIDIA RTX 5090 GPU.
For the initial $6$-step rollout stage, one training epoch takes approximately $16$ minutes for the single-object setting and increases with the number of interacting objects in the two-, three-, and five-object mixed-contact settings.

\paragraph{Model capacity.}
Unless otherwise specified, learned models use hidden dimension $128$, MLP depth $2$, and $4$ message-passing steps.
We adjust message-passing depth for standard baseline conventions: Interaction Network uses $2$ message steps, the GNS-style model uses $6$ message steps, and the Point Transformer acceleration baseline uses $2$ message steps.
Most other baselines use $4$ message steps.

\paragraph{Rollout settings.}
All baselines use the same graph construction procedure and the same rollout substep count of $4$.
The rollout loss uses $\lambda_x=1.0$ for position and $\lambda_v=0.25$ for velocity, matching the main training objective.
Single-object results are averaged over three random seeds. Multi-object results are evaluated using one best-validation checkpoint per method and object count.

\begin{table}[H]
\centering
\caption{Three-seed uncertainty for the primary long-horizon rollout metrics. We report mean $\pm$ standard deviation over random seeds.}
\label{tab:seed-uncertainty}
\small
\resizebox{0.7\textwidth}{!}{%
\begin{tabular}{lcc}
\toprule
Model / Variant & RE@24 & RE@48 \\
\midrule
Full \method & \textbf{0.0873 $\pm$ 0.0033} & \textbf{0.1367 $\pm$ 0.0061} \\
No dissipation & 0.0947 $\pm$ 0.0031 & 0.1536 $\pm$ 0.0216 \\
No extra force & 0.1024 $\pm$ 0.0018 & 0.2177 $\pm$ 0.0038 \\
Local graph only & 0.1139 $\pm$ 0.0031 & 0.1731 $\pm$ 0.0038 \\
Direct-Acceleration GNN & 0.1237 $\pm$ 0.0027 & 0.2169 $\pm$ 0.0021 \\
EGNN Dynamics & 0.1239 $\pm$ 0.0026 & 0.2188 $\pm$ 0.0042 \\
Graph Network Simulator (GNS) & 0.1244 $\pm$ 0.0010 & 0.2126 $\pm$ 0.0019 \\
Interaction Network & 0.1250 $\pm$ 0.0007 & 0.2190 $\pm$ 0.0019 \\
Direct-State GNN & 0.1350 $\pm$ 0.0052 & 0.2315 $\pm$ 0.0181 \\
Point Transformer Dynamics & 0.1367 $\pm$ 0.0017 & 0.2274 $\pm$ 0.0020 \\
Explicit KNN Energy & 0.1601 $\pm$ 0.0038 & 0.3062 $\pm$ 0.0028 \\
No stiffness $k$ & 0.1899 $\pm$ 0.0005 & 0.1769 $\pm$ 0.0026 \\
No structural energy & 0.2351 $\pm$ 0.0031 & 0.3532 $\pm$ 0.0010 \\
Point-point contact & 0.8643 $\pm$ 0.0387 & 3.6664 $\pm$ 0.1934 \\
No contact & 1.6834 $\pm$ 0.0001 & 7.3486 $\pm$ 0.0004 \\
\bottomrule
\end{tabular}}
\end{table}

\subsection{Stiffness-Conditioned Results}
\label{app:stiffness-results}

Table~\ref{tab:by-k} breaks down the main RE@24 metric by stiffness value for representative methods and the no-stiffness ablation.

\begin{table}[H]
\centering
\caption{RE@24 by stiffness value.}
\label{tab:by-k}
\small
\begin{tabular}{lcccccc}
\toprule
Model & $k=10$ & $20$ & $50$ & $100$ & $200$ & $500$ \\
\midrule
\method & \textbf{0.078} & \textbf{0.083} & \textbf{0.079} & \textbf{0.077} & \textbf{0.093} & \textbf{0.116} \\
Direct-Acceleration GNN & 0.110 & 0.121 & 0.120 & 0.126 & 0.131 & 0.135 \\
Graph Network Simulator (GNS) & 0.105 & 0.119 & 0.125 & 0.129 & 0.133 & 0.139 \\
Point Transformer Dynamics & 0.130 & 0.140 & 0.140 & 0.141 & 0.138 & 0.141 \\
No stiffness $k$ & 0.197 & 0.133 & 0.136 & 0.181 & 0.229 & 0.262 \\
\bottomrule
\end{tabular}
\end{table}

\subsection{Video Generation Implementation}
\label{app:video-details}

We use \method\ as a trajectory-generation module upstream of a trajectory-conditioned image-to-video generator.
The purpose of this experiment is to test whether improved physical rollout quality transfers to video synthesis through explicit motion control.

\paragraph{Pipeline.}
Given an input image, we estimate an object mask, reconstruct an object point cloud and an environment point cloud, and apply user-specified physical controls.
\method\ rolls out 3D object trajectories under gravity, stiffness, contact, and optional external forces.
The resulting 3D trajectories are projected to the image plane and converted into dense 2D track maps used by the video generator.

\paragraph{Compared methods.}
We compare with text/image-conditioned and motion-controlled video-generation baselines: Wan2.2~\citep{wan2025}, CogVideoX-5B~\citep{yang2024cogvideox}, DragAnything~\citep{wu2024draganything}, ObjCtrl-2.5D~\citep{wang2024objctrl}, Veo3~\citep{googledeepmind2026veo}, and PhysCtrl~\citep{2025arXiv250920358W}.
For methods that accept only sparse motion control, we provide a centroid or single-point trajectory according to the method interface.
For \method, we use dense projected multi-point trajectories from the physics rollout.

\paragraph{Test cases.}
The main paper reports results on six contact-rich cases: car, cardboard box, duck, football, pig, and sponge.
The prompts are:
\begin{itemize}[leftmargin=*]
    \item \textbf{Car:} A miniature red toy SUV falls from the air, hits the wooden floor, bounces, slides forward, and finally settles, in a warm cozy living room with soft depth of field, realistic collision and motion.
    \item \textbf{Cardboard box:} A cardboard box drops onto the table, collides, bounces slightly, and then settles with realistic motion.
    \item \textbf{Duck:} A toy duck drops onto the bathroom floor, collides, bounces, and settles with realistic plastic toy motion.
    \item \textbf{Football:} A football drops onto the grass, collides, rebounds, and then rolls to a stop with realistic motion.
    \item \textbf{Pig:} A toy pig falls to the bedroom floor, collides, rebounds slightly, and comes to rest with realistic motion.
    \item \textbf{Sponge:} A sponge-like object drops onto the floor, collides, and settles with realistic elastic motion.
\end{itemize}

\paragraph{Evaluation protocol.}
We use GPT-5.5 as a 5-point Likert evaluator over Semantic Adherence (SA), Physical Commonsense (PC), and Video Quality (VQ), with evaluations conducted on May 1, 2026.
For each method and case, the evaluator receives the full generated video as primary evidence and ten uniformly sampled frames as reference.
The evaluator is instructed to use full-video motion as the primary evidence, use the frame grid only as reference, apply the same standard across methods, and not favor methods by name.
The evaluator prompt is:
\begin{quote}
\small
You are evaluating physics-grounded video generation.

For this test case, I provide:
1. The case prompt (scene + physical interaction intent).
2. Method videos (one video per method) in the same folder.
3. A 10-frame reference grid image (\texttt{combined\_grid.jpg}).

Please score each method on a 5-point Likert scale:
1 = poor, 2 = fair, 3 = acceptable, 4 = good, 5 = excellent.

Criteria:
Semantic Adherence (SA): whether generated content and motion match the text prompt, input scene, object identity, physical condition, and intended interaction.
Physical Commonsense (PC): whether motion is physically plausible, including force response, contact, rebound, damping, stiffness behavior, and absence of severe penetration/floating/teleportation.
Video Quality (VQ): visual quality and temporal consistency. Penalize flicker, disappearance, identity drift, severe artifacts, unrealistic hallucinations, broken geometry, and static motion when motion is expected.

Important:
Use full video motion as primary evidence.
Use \texttt{combined\_grid.jpg} only as reference.
Use the same standard across all methods.
Do not favor methods by name.
A visually good video can still receive low PC if physics is implausible.

\texttt{case\_id}: [case id]

Prompt: [case prompt]

\end{quote}

\paragraph{Human evaluation.}
To complement the GPT-based scores, we conducted a blind human evaluation with $10$ non-author volunteer raters.
The raters used the same case prompts, full method videos, $10$-frame reference grids, and 5-point SA/PC/VQ scoring criteria as the GPT evaluator.
Method names were anonymized and presentation order was randomized.
Raters were informed that the task was an anonymous research evaluation of synthetic object videos, participation was voluntary, and responses would be reported only in aggregate.
No personal, sensitive, or biometric data was collected from raters, and no compensation was provided.
The aggregate human scores averaged over cases and raters are reported together with the GPT-based scores in Table~\ref{tab:video-eval}.

\paragraph{Generation settings.}
All generated videos are evaluated at $720\times480$ resolution, $24$ FPS, and $48$ frames.

\section*{Reproducibility Statement}

We provide the data-generation protocol, train/test split specification, force schedules, stiffness settings, model hyperparameters, and baseline training configurations.
The appendix reports dataset statistics, rollout protocols, implementation constants, and video-evaluation settings needed to reproduce the main tables and qualitative comparisons.
The multi-object NEXUS implementation and instructions for data generation, training, and evaluation are available at \url{https://github.com/StephenYing/NEXUS}.

\section*{Broader Impact}

Structured physical trajectory models can benefit robotics simulation, physically grounded animation, scientific visualization, and controllable media generation by making generated motion more physically plausible and easier to control.
At the same time, improved physical realism can make synthetic videos more convincing, which may amplify risks of misleading or misattributed generated content.
When used for media synthesis, generated videos should be labeled as synthetic, provenance metadata should be preserved when possible, and downstream deployments should respect consent, copyright, and platform disclosure policies.
Our work focuses on contact-rich object dynamics rather than human identity, speech, or biometric generation, but the same transparency practices remain important for responsible use.

\newpage
\section*{NeurIPS Paper Checklist}

\begin{enumerate}

\item {\bf Claims}
    \item[] Question: Do the main claims made in the abstract and introduction accurately reflect the paper's contributions and scope?
    \item[] Answer: \answerYes{}
    \item[] Justification: The main claims and contributions are stated in the Abstract and Introduction, and the experimental scope is reflected in the Experiments and Discussion and Limitations sections.
    \item[] Guidelines:
    \begin{itemize}
        \item The answer \answerNA{} means that the abstract and introduction do not include the claims made in the paper.
        \item The abstract and/or introduction should clearly state the claims made, including the contributions made in the paper and important assumptions and limitations. A \answerNo{} or \answerNA{} answer to this question will not be perceived well by the reviewers. 
        \item The claims made should match theoretical and experimental results, and reflect how much the results can be expected to generalize to other settings. 
        \item It is fine to include aspirational goals as motivation as long as it is clear that these goals are not attained by the paper. 
    \end{itemize}

\item {\bf Limitations}
    \item[] Question: Does the paper discuss the limitations of the work performed by the authors?
    \item[] Answer: \answerYes{}
    \item[] Justification: Section 5 discusses the main limitations, including the controlled object--environment benchmark scope, the multi-object stress-test scope, metric trade-offs, and future extensions.
    \item[] Guidelines:
    \begin{itemize}
        \item The answer \answerNA{} means that the paper has no limitation while the answer \answerNo{} means that the paper has limitations, but those are not discussed in the paper. 
        \item The authors are encouraged to create a separate ``Limitations'' section in their paper.
        \item The paper should point out any strong assumptions and how robust the results are to violations of these assumptions (e.g., independence assumptions, noiseless settings, model well-specification, asymptotic approximations only holding locally). The authors should reflect on how these assumptions might be violated in practice and what the implications would be.
        \item The authors should reflect on the scope of the claims made, e.g., if the approach was only tested on a few datasets or with a few runs. In general, empirical results often depend on implicit assumptions, which should be articulated.
        \item The authors should reflect on the factors that influence the performance of the approach. For example, a facial recognition algorithm may perform poorly when image resolution is low or images are taken in low lighting. Or a speech-to-text system might not be used reliably to provide closed captions for online lectures because it fails to handle technical jargon.
        \item The authors should discuss the computational efficiency of the proposed algorithms and how they scale with dataset size.
        \item If applicable, the authors should discuss possible limitations of their approach to address problems of privacy and fairness.
        \item While the authors might fear that complete honesty about limitations might be used by reviewers as grounds for rejection, a worse outcome might be that reviewers discover limitations that aren't acknowledged in the paper. The authors should use their best judgment and recognize that individual actions in favor of transparency play an important role in developing norms that preserve the integrity of the community. Reviewers will be specifically instructed to not penalize honesty concerning limitations.
    \end{itemize}

\item {\bf Theory assumptions and proofs}
    \item[] Question: For each theoretical result, does the paper provide the full set of assumptions and a complete (and correct) proof?
    \item[] Answer: \answerNA{}
    \item[] Justification: The paper introduces a modeling formulation and empirical evaluation, but does not claim new theoretical theorems or proofs.
    \item[] Guidelines:
    \begin{itemize}
        \item The answer \answerNA{} means that the paper does not include theoretical results. 
        \item All the theorems, formulas, and proofs in the paper should be numbered and cross-referenced.
        \item All assumptions should be clearly stated or referenced in the statement of any theorems.
        \item The proofs can either appear in the main paper or the supplemental material, but if they appear in the supplemental material, the authors are encouraged to provide a short proof sketch to provide intuition. 
        \item Inversely, any informal proof provided in the core of the paper should be complemented by formal proofs provided in appendix or supplemental material.
        \item Theorems and Lemmas that the proof relies upon should be properly referenced. 
    \end{itemize}

    \item {\bf Experimental result reproducibility}
    \item[] Question: Does the paper fully disclose all the information needed to reproduce the main experimental results of the paper to the extent that it affects the main claims and/or conclusions of the paper (regardless of whether the code and data are provided or not)?
    \item[] Answer: \answerYes{}
    \item[] Justification: Section 4 describes the benchmark, metrics, baselines, ablations, and video-evaluation protocol; the appendix reports dataset statistics, graph construction, training details, and baseline settings. The multi-object NEXUS code and usage instructions are available at \url{https://github.com/StephenYing/NEXUS}.
    \item[] Guidelines:
    \begin{itemize}
        \item The answer \answerNA{} means that the paper does not include experiments.
        \item If the paper includes experiments, a \answerNo{} answer to this question will not be perceived well by the reviewers: Making the paper reproducible is important, regardless of whether the code and data are provided or not.
        \item If the contribution is a dataset and\slash or model, the authors should describe the steps taken to make their results reproducible or verifiable. 
        \item Depending on the contribution, reproducibility can be accomplished in various ways. For example, if the contribution is a novel architecture, describing the architecture fully might suffice, or if the contribution is a specific model and empirical evaluation, it may be necessary to either make it possible for others to replicate the model with the same dataset, or provide access to the model. In general. releasing code and data is often one good way to accomplish this, but reproducibility can also be provided via detailed instructions for how to replicate the results, access to a hosted model (e.g., in the case of a large language model), releasing of a model checkpoint, or other means that are appropriate to the research performed.
        \item While NeurIPS does not require releasing code, the conference does require all submissions to provide some reasonable avenue for reproducibility, which may depend on the nature of the contribution. For example
        \begin{enumerate}
            \item If the contribution is primarily a new algorithm, the paper should make it clear how to reproduce that algorithm.
            \item If the contribution is primarily a new model architecture, the paper should describe the architecture clearly and fully.
            \item If the contribution is a new model (e.g., a large language model), then there should either be a way to access this model for reproducing the results or a way to reproduce the model (e.g., with an open-source dataset or instructions for how to construct the dataset).
            \item We recognize that reproducibility may be tricky in some cases, in which case authors are welcome to describe the particular way they provide for reproducibility. In the case of closed-source models, it may be that access to the model is limited in some way (e.g., to registered users), but it should be possible for other researchers to have some path to reproducing or verifying the results.
        \end{enumerate}
    \end{itemize}

\item {\bf Open access to data and code}
    \item[] Question: Does the paper provide open access to the data and code, with sufficient instructions to faithfully reproduce the main experimental results, as described in supplemental material?
    \item[] Answer: \answerYes{}
    \item[] Justification: The multi-object NEXUS implementation and instructions for data generation, training, and evaluation are available at \url{https://github.com/StephenYing/NEXUS}.
    \item[] Guidelines:
    \begin{itemize}
        \item The answer \answerNA{} means that paper does not include experiments requiring code.
        \item Please see the NeurIPS code and data submission guidelines (\url{https://neurips.cc/public/guides/CodeSubmissionPolicy}) for more details.
        \item While we encourage the release of code and data, we understand that this might not be possible, so \answerNo{} is an acceptable answer. Papers cannot be rejected simply for not including code, unless this is central to the contribution (e.g., for a new open-source benchmark).
        \item The instructions should contain the exact command and environment needed to run to reproduce the results. See the NeurIPS code and data submission guidelines (\url{https://neurips.cc/public/guides/CodeSubmissionPolicy}) for more details.
        \item The authors should provide instructions on data access and preparation, including how to access the raw data, preprocessed data, intermediate data, and generated data, etc.
        \item The authors should provide scripts to reproduce all experimental results for the new proposed method and baselines. If only a subset of experiments are reproducible, they should state which ones are omitted from the script and why.
        \item At submission time, to preserve anonymity, the authors should release anonymized versions (if applicable).
        \item Providing as much information as possible in supplemental material (appended to the paper) is recommended, but including URLs to data and code is permitted.
    \end{itemize}

\item {\bf Experimental setting/details}
    \item[] Question: Does the paper specify all the training and test details (e.g., data splits, hyperparameters, how they were chosen, type of optimizer) necessary to understand the results?
    \item[] Answer: \answerYes{}
    \item[] Justification: The Experiments section and appendix specify data splits, rollout horizons, metrics, graph settings, optimizer, training curriculum, model capacity, and baseline configurations.
    \item[] Guidelines:
    \begin{itemize}
        \item The answer \answerNA{} means that the paper does not include experiments.
        \item The experimental setting should be presented in the core of the paper to a level of detail that is necessary to appreciate the results and make sense of them.
        \item The full details can be provided either with the code, in appendix, or as supplemental material.
    \end{itemize}

\item {\bf Experiment statistical significance}
    \item[] Question: Does the paper report error bars suitably and correctly defined or other appropriate information about the statistical significance of the experiments?
    \item[] Answer: \answerYes{}
    \item[] Justification: The main dynamics results are averaged over three random seeds, and Appendix Table~\ref{tab:seed-uncertainty} reports mean and standard deviation for the primary RE@24 and RE@48 metrics.
    \item[] Guidelines:
    \begin{itemize}
        \item The answer \answerNA{} means that the paper does not include experiments.
        \item The authors should answer \answerYes{} if the results are accompanied by error bars, confidence intervals, or statistical significance tests, at least for the experiments that support the main claims of the paper.
        \item The factors of variability that the error bars are capturing should be clearly stated (for example, train/test split, initialization, random drawing of some parameter, or overall run with given experimental conditions).
        \item The method for calculating the error bars should be explained (closed form formula, call to a library function, bootstrap, etc.)
        \item The assumptions made should be given (e.g., Normally distributed errors).
        \item It should be clear whether the error bar is the standard deviation or the standard error of the mean.
        \item It is OK to report 1-sigma error bars, but one should state it. The authors should preferably report a 2-sigma error bar than state that they have a 96\% CI, if the hypothesis of Normality of errors is not verified.
        \item For asymmetric distributions, the authors should be careful not to show in tables or figures symmetric error bars that would yield results that are out of range (e.g., negative error rates).
        \item If error bars are reported in tables or plots, the authors should explain in the text how they were calculated and reference the corresponding figures or tables in the text.
    \end{itemize}

\item {\bf Experiments compute resources}
    \item[] Question: For each experiment, does the paper provide sufficient information on the computer resources (type of compute workers, memory, time of execution) needed to reproduce the experiments?
    \item[] Answer: \answerYes{}
    \item[] Justification: The appendix reports the GPU type, number of epochs, rollout windows, batch size, training schedule, and representative per-epoch wall-clock training time for the single-, two-, and three-object settings.
    \item[] Guidelines:
    \begin{itemize}
        \item The answer \answerNA{} means that the paper does not include experiments.
        \item The paper should indicate the type of compute workers CPU or GPU, internal cluster, or cloud provider, including relevant memory and storage.
        \item The paper should provide the amount of compute required for each of the individual experimental runs as well as estimate the total compute. 
        \item The paper should disclose whether the full research project required more compute than the experiments reported in the paper (e.g., preliminary or failed experiments that didn't make it into the paper). 
    \end{itemize}
    
\item {\bf Code of ethics}
    \item[] Question: Does the research conducted in the paper conform, in every respect, with the NeurIPS Code of Ethics \url{https://neurips.cc/public/EthicsGuidelines}?
    \item[] Answer: \answerYes{}
    \item[] Justification: The work uses standard simulation, benchmark evaluation, and synthetic video-generation protocols, and we are not aware of any deviation from the NeurIPS Code of Ethics.
    \item[] Guidelines:
    \begin{itemize}
        \item The answer \answerNA{} means that the authors have not reviewed the NeurIPS Code of Ethics.
        \item If the authors answer \answerNo, they should explain the special circumstances that require a deviation from the Code of Ethics.
        \item The authors should make sure to preserve anonymity (e.g., if there is a special consideration due to laws or regulations in their jurisdiction).
    \end{itemize}

\item {\bf Broader impacts}
    \item[] Question: Does the paper discuss both potential positive societal impacts and negative societal impacts of the work performed?
    \item[] Answer: \answerYes{}
    \item[] Justification: The Broader Impact section discusses positive applications in simulation, animation, visualization, and controllable media, as well as risks from more realistic synthetic videos and suggested transparency practices.
    \item[] Guidelines:
    \begin{itemize}
        \item The answer \answerNA{} means that there is no societal impact of the work performed.
        \item If the authors answer \answerNA{} or \answerNo, they should explain why their work has no societal impact or why the paper does not address societal impact.
        \item Examples of negative societal impacts include potential malicious or unintended uses (e.g., disinformation, generating fake profiles, surveillance), fairness considerations (e.g., deployment of technologies that could make decisions that unfairly impact specific groups), privacy considerations, and security considerations.
        \item The conference expects that many papers will be foundational research and not tied to particular applications, let alone deployments. However, if there is a direct path to any negative applications, the authors should point it out. For example, it is legitimate to point out that an improvement in the quality of generative models could be used to generate Deepfakes for disinformation. On the other hand, it is not needed to point out that a generic algorithm for optimizing neural networks could enable people to train models that generate Deepfakes faster.
        \item The authors should consider possible harms that could arise when the technology is being used as intended and functioning correctly, harms that could arise when the technology is being used as intended but gives incorrect results, and harms following from (intentional or unintentional) misuse of the technology.
        \item If there are negative societal impacts, the authors could also discuss possible mitigation strategies (e.g., gated release of models, providing defenses in addition to attacks, mechanisms for monitoring misuse, mechanisms to monitor how a system learns from feedback over time, improving the efficiency and accessibility of ML).
    \end{itemize}
    
\item {\bf Safeguards}
    \item[] Question: Does the paper describe safeguards that have been put in place for responsible release of data or models that have a high risk for misuse (e.g., pre-trained language models, image generators, or scraped datasets)?
    \item[] Answer: \answerNA{}
    \item[] Justification: The paper concerns a dynamics model rather than a high-risk pretrained language, image, or video generation model; broader synthetic-media risks are discussed in the Broader Impact section.
    \item[] Guidelines:
    \begin{itemize}
        \item The answer \answerNA{} means that the paper poses no such risks.
        \item Released models that have a high risk for misuse or dual-use should be released with necessary safeguards to allow for controlled use of the model, for example by requiring that users adhere to usage guidelines or restrictions to access the model or implementing safety filters. 
        \item Datasets that have been scraped from the Internet could pose safety risks. The authors should describe how they avoided releasing unsafe images.
        \item We recognize that providing effective safeguards is challenging, and many papers do not require this, but we encourage authors to take this into account and make a best faith effort.
    \end{itemize}

\item {\bf Licenses for existing assets}
    \item[] Question: Are the creators or original owners of assets (e.g., code, data, models), used in the paper, properly credited and are the license and terms of use explicitly mentioned and properly respected?
    \item[] Answer: \answerYes{}
    \item[] Justification: Existing datasets, baselines, and pretrained video models used for comparison or downstream evaluation are cited in the paper, and the supplementary release will follow their respective licenses and terms of use.
    \item[] Guidelines:
    \begin{itemize}
        \item The answer \answerNA{} means that the paper does not use existing assets.
        \item The authors should cite the original paper that produced the code package or dataset.
        \item The authors should state which version of the asset is used and, if possible, include a URL.
        \item The name of the license (e.g., CC-BY 4.0) should be included for each asset.
        \item For scraped data from a particular source (e.g., website), the copyright and terms of service of that source should be provided.
        \item If assets are released, the license, copyright information, and terms of use in the package should be provided. For popular datasets, \url{paperswithcode.com/datasets} has curated licenses for some datasets. Their licensing guide can help determine the license of a dataset.
        \item For existing datasets that are re-packaged, both the original license and the license of the derived asset (if it has changed) should be provided.
        \item If this information is not available online, the authors are encouraged to reach out to the asset's creators.
    \end{itemize}

\item {\bf New assets}
    \item[] Question: Are new assets introduced in the paper well documented and is the documentation provided alongside the assets?
    \item[] Answer: \answerNo{}
    \item[] Justification: The paper introduces a new model and benchmark data, but the code, model, checkpoints, and associated documentation will be released after acceptance rather than during submission.
    \item[] Guidelines:
    \begin{itemize}
        \item The answer \answerNA{} means that the paper does not release new assets.
        \item Researchers should communicate the details of the dataset\slash code\slash model as part of their submissions via structured templates. This includes details about training, license, limitations, etc. 
        \item The paper should discuss whether and how consent was obtained from people whose asset is used.
        \item At submission time, remember to anonymize your assets (if applicable). You can either create an anonymized URL or include an anonymized zip file.
    \end{itemize}

\item {\bf Crowdsourcing and research with human subjects}
    \item[] Question: For crowdsourcing experiments and research with human subjects, does the paper include the full text of instructions given to participants and screenshots, if applicable, as well as details about compensation (if any)? 
    \item[] Answer: \answerYes{}
    \item[] Justification: The appendix describes the blind human video evaluation protocol. We used the same prompt and 5-point SA/PC/VQ scoring instructions as the GPT evaluator, with ten non-author volunteer raters, anonymized method names, randomized presentation order, voluntary participation, aggregate reporting, no compensation, and no personal data collection.
    \item[] Guidelines:
    \begin{itemize}
        \item The answer \answerNA{} means that the paper does not involve crowdsourcing nor research with human subjects.
        \item Including this information in the supplemental material is fine, but if the main contribution of the paper involves human subjects, then as much detail as possible should be included in the main paper. 
        \item According to the NeurIPS Code of Ethics, workers involved in data collection, curation, or other labor should be paid at least the minimum wage in the country of the data collector. 
    \end{itemize}

\item {\bf Institutional review board (IRB) approvals or equivalent for research with human subjects}
    \item[] Question: Does the paper describe potential risks incurred by study participants, whether such risks were disclosed to the subjects, and whether Institutional Review Board (IRB) approvals (or an equivalent approval/review based on the requirements of your country or institution) were obtained?
    \item[] Answer: \answerYes{}
    \item[] Justification: The appendix describes the human evaluation as an anonymous, voluntary, minimal-risk rating of generated object videos. Raters were informed of the task, responses were reported only in aggregate, no personal, sensitive, or biometric data was collected, and no identifying institutional review information is included in the anonymous submission.
    \item[] Guidelines:
    \begin{itemize}
        \item The answer \answerNA{} means that the paper does not involve crowdsourcing nor research with human subjects.
        \item Depending on the country in which research is conducted, IRB approval (or equivalent) may be required for any human subjects research. If you obtained IRB approval, you should clearly state this in the paper. 
        \item We recognize that the procedures for this may vary significantly between institutions and locations, and we expect authors to adhere to the NeurIPS Code of Ethics and the guidelines for their institution. 
        \item For initial submissions, do not include any information that would break anonymity (if applicable), such as the institution conducting the review.
    \end{itemize}

\item {\bf Declaration of LLM usage}
    \item[] Question: Does the paper describe the usage of LLMs if it is an important, original, or non-standard component of the core methods in this research? Note that if the LLM is used only for writing, editing, or formatting purposes and does \emph{not} impact the core methodology, scientific rigor, or originality of the research, declaration is not required.
    \item[] Answer: \answerNA{}
    \item[] Justification: LLMs are not used as an important, original, or non-standard component of the proposed method; any use is limited to writing, editing, or formatting support.
    \item[] Guidelines:
    \begin{itemize}
        \item The answer \answerNA{} means that the core method development in this research does not involve LLMs as any important, original, or non-standard components.
        \item Please refer to our LLM policy in the NeurIPS handbook for what should or should not be described.
    \end{itemize}

\end{enumerate}

\end{document}